\documentclass[10pt,twocolumn,letterpaper]{article}

\usepackage{cvpr}              

\usepackage{graphicx}
\usepackage{amsmath}
\usepackage{amssymb}
\usepackage{booktabs}
\usepackage{multirow}
\usepackage[utf8]{inputenc}
\usepackage[table,xcdraw]{xcolor}
\usepackage[mathscr]{euscript}
\usepackage[accsupp]{axessibility}

\usepackage[pagebackref,breaklinks,colorlinks]{hyperref}

\usepackage{soul}
\usepackage[capitalize]{cleveref}
\crefname{section}{Sec.}{Secs.}
\Crefname{section}{Section}{Sections}
\Crefname{table}{Table}{Tables}
\crefname{table}{Tab.}{Tabs.}

\def\cvprPaperID{10254} 
\def\confName{CVPR}
\def\confYear{2023}

\begin{document}

\title{METransformer: Radiology Report Generation by Transformer with Multiple Learnable Expert Tokens}

\author{Zhanyu Wang\\
University of Sydney\\
{\tt\small zhanyu.wang@sydney.edu.au}
\and
Lingqiao Liu\\
University of Adelaide\\
{\tt\small lingqiao.liu@adelaide.edu.au}
\and
Lei Wang\\
University of Wollongong\\
{\tt\small leiw@uow.edu.au}
\and
Luping Zhou\\
University of Sydney\\
{\tt\small luping.zhou@sydney.edu.au}
}
\maketitle

\begin{abstract}
In clinical scenarios, multi-specialist consultation could significantly benefit the diagnosis, especially for intricate cases. This inspires us to explore a ``multi-expert joint diagnosis" mechanism to upgrade the existing “single expert” framework commonly seen in the current literature. To this end, we propose METransformer, a method to realize this idea with a transformer-based backbone. The key design of our method is the introduction of multiple learnable ``expert” tokens into both the transformer encoder and decoder. In the encoder, each expert token interacts with both vision tokens and other expert tokens to learn to attend different image regions for image representation. These expert tokens are encouraged to capture complementary information by an orthogonal loss that minimizes their overlap. In the decoder, each attended expert token guides the cross-attention between input words and visual tokens, thus influencing the generated report. A metrics-based expert voting strategy is further developed to generate the final report. By the multi-experts concept, our model enjoys the merits of an ensemble-based approach but through a manner that is computationally more efficient and supports more sophisticated interactions among experts. Experimental results demonstrate the promising performance of our proposed model on two widely used benchmarks. Last but not least, the framework-level innovation makes our work ready to incorporate advances on existing ``single-expert" models to further improve its performance.

\end{abstract}

\section{Introduction}
\label{sec:intro}

Interpreting radiology images (e.g., chest X-ray) and writing diagnostic reports are essential operations in clinical practice and normally require a considerable manual workload. Therefore, radiology report generation, which aims to automatically generate a free-text description based on a radiograph, is highly desired to ease the burden of radiologists while maintaining the quality of health care. Recently, substantial progress has been made towards research on automated radiology report generation models~\cite{li2018hybrid, Knowledge-Driven, 2020When,2020icdmAutomatic ,chen2020generating, li2020auxiliary,Wang_2021_CVPR, CVPR201_PPKD, chen2022cross, 2021Knowledge ,wang2022medical,2022AlignTransformer,2021Auto,wang2022tmi}. Most existing studies adopt a conventional encoder-decoder architecture following the image captioning paradigm~\cite{vinyals2015showandtell,xu2016showattandtell,2020Meshed,zhou2020moregrounded,pan2020xlinear} and resort to optimizing network structure or introducing external or prior information to aid report generation. These methods, in this paper, are collectively referred to as ``single-expert" based diagnostic captioning methods.

However, diagnostic report generation is a very challenging task as disease anomalies usually only occupy a small portion of the whole image and could appear at arbitrary locations. Due to the fine-grained nature of radiology images, it is hard to focus on the correct image regions throughout the report generation procedure despite different attentions developed in recent works~\cite{li2020auxiliary, CVPR201_PPKD}. Meanwhile, it is noticed that in clinic scenarios, multi-specialist consultation is especially beneficial for those intricate diagnostic cases that challenge a single specialist for a comprehensive and accurate diagnosis. The above observations led us to think, could we design a model to simulate the  multi-specialist consultation scenario? Based on this motivation, we propose a new diagnostic captioning framework, METransformer, to mimic the ``multi-expert joint diagnosis" process. Built upon a transformer backbone, METransformer introduces multiple ``expert tokens", representing multiple experts, into both the transformer encoder and decoder. Each expert token learns to attend distinct image regions and interacts with other expert tokens to capture reliable and complementary visual information and produce a diagnosis report in parallel. The optimal report is selected through an expert voting strategy to produce the final report.   Our design is based on the assumption that it would be easier for multiple experts than a single one to capture visual patterns of clinical importance, which is verified by our experimental results.

Specifically, we feed both the expert tokens (learnable embeddings) and the visual tokens  (image patches embeddings) into the Expert Transformer encoder which is comprised of a vision transformer (ViT) encoder and a bilinear transformer encoder.  In ViT encoder, each expert token interacts not only with the visual tokens but also with the other expert tokens. Further, in the bilinear transformer encoder, to enable each ``expert" to capture fine-grained image information, we compute higher-order attention between expert tokens and visual tokens, which has proved to be effective in fine-grained classification tasks~\cite{lin2015bilinear}. It is noteworthy that the expert tokens in the encoder are encouraged to learn complementary representations by an orthogonal loss so that they attend differently to different image regions.  With these carefully learned expert token embeddings, in the decoder, we take them as a guide to regulate the learning of word embeddings and visual token embedding in the report generation process. This results in M different word and visual token embeddings, thus  producing M candidate reports, where M is the number of experts. We further propose a metric-based expert voting strategy to generate the final report from the M candidates.

By utilizing multiple experts, our model, to some extent, is analogous to ensemble-based approaches, while each expert token corresponds to an individual model. While it enjoys the merits of ensemble-based approaches, our model is designed in a manner that is computationally more efficient and supports more sophisticated interactions among experts. Therefore, it can scale up with only a trivial increase of model parameters and achieves better performance, as demonstrated in our experimental study.

Our main contributions are summarized as follows.

First, we propose a new diagnostic captioning framework, METransformer,  which is conceptually ``multi-expert joint diagnosis" for radiology report generation, by introducing learnable expert tokens and encouraging them to learn complementary representations using both linear and non-linear attentions.

Second, our model enjoys the benefits of an ensemble approach. Thanks to the carefully designed network structure and the end-to-end training manner, our model can achieve better results than common ensemble approaches while greatly reducing training parameters and improving training efficiency.

Third, our approach shows promising performance on two widely used benchmarks IU-Xray and MIMIC-CXR over multiple state-of-the-art methods.  The clinic relevance of the generated reports is also analyzed.
\section{Related Work}
\label{sec:relate}
\noindent \textbf{Image Captioning.}~~Natural image captioning task aims at automatically generating a single sentence description for the given image. A broad collection of methods was proposed in the last few years~\cite{vinyals2015showandtell, xu2016showattandtell,2016semanticattention,2016reviewnetwork,2017Knowing, rennie2017selfcritical,topdown,2020Meshed,zhou2020moregrounded, pan2020xlinear} and have achieved great success in advancing the state-of-the-art.  Most of them adopt the conventional encoder-decoder architecture, with convolutional neural networks (CNNs) as the encoder and recurrent (e.g., LSTM/GRU) or non-recurrent networks (e.g., Transformer) as the decoder with a carefully designed attention module~\cite{xu2016showattandtell, 2017Knowing,pan2020xlinear} to produce image description. However, compare with image captioning, radiographic images have fine-grained characteristics and radiology report generation aims to generate a long paragraph rather than a single sentence, which brings more challenge to the model's attention ability.
\\
\noindent \textbf{Medical report generation.}~~Most of the research efforts in medical report generation can be roughly categorized as being along two main directions.  The first direction lies in promoting the model's structure, such as introducing a better attention mechanism or improving the structure of the report decoder.  For example, some works~\cite{2017Ontheautomatic, 2018Multimodal, 2019ipmiImproved, 2019muiti-view, 2020icdmAutomatic, Wang_2021_CVPR} utilize a hierarchically structured LSTM network 
to better handle the long narrative nature of medical reports.   Jing et al~\cite{2017Ontheautomatic} proposed a multi-task hierarchical model with co-attention by automatically predicting keywords to assist in generating long paragraphs. Xue et al~\cite{2018Multimodal, 2019ipmiImproved} presented a different network structure involving a generative sentence model and a generative paragraph model that uses the generated sentence to produce the next sentence. In addition, Wang et al~\cite{Wang_2021_CVPR} introduced an image-report matching network to bridge the domain gap between image and text for reducing the difficulty of report generation.  To further improve performance, some works~\cite{chen2020generating, chen2022cross,wang2022tmi} employ a transformer instead of LSTM as the report decoder, which has achieved good results. Work~\cite{chen2020generating} proposes to generate radiographic reports with a memory-driven Transformer designed to record key information of the generation process.  The second research direction studies how to leverage medical domain knowledge to guide report generation. Most recently, many works~\cite{2020When,li2020auxiliary,2021Radiology, 2021Knowledge, CVPR201_PPKD, 2021Auto} attempt to integrate knowledge graphs into medical report generation pipeline to improve the quality of the generated reports. Another group of works~\cite{2022AlignTransformer, wang2022medical} utilizes disease tags to facilitate report generation. Yang et al~\cite{2021Knowledge} present a framework based on both general and specific knowledge, where the general knowledge is obtained from a pre-constructed knowledge graph, while the specific knowledge is derived from retrieving similar reports. The work~\cite{wang2022medical} proposed a medical concepts generation network to generate semantic information and integrate it into the report generation process. It is worth mentioning that our approach is orthogonal to the methods mentioned above, for example, the advanced memory-enhanced decoder used in ~\cite{chen2020generating} can also be applied to our framework for further performance improvement.

\section{Method}
\begin{figure*}[t]
    \centering
    \centerline{\includegraphics[width=0.95\linewidth]{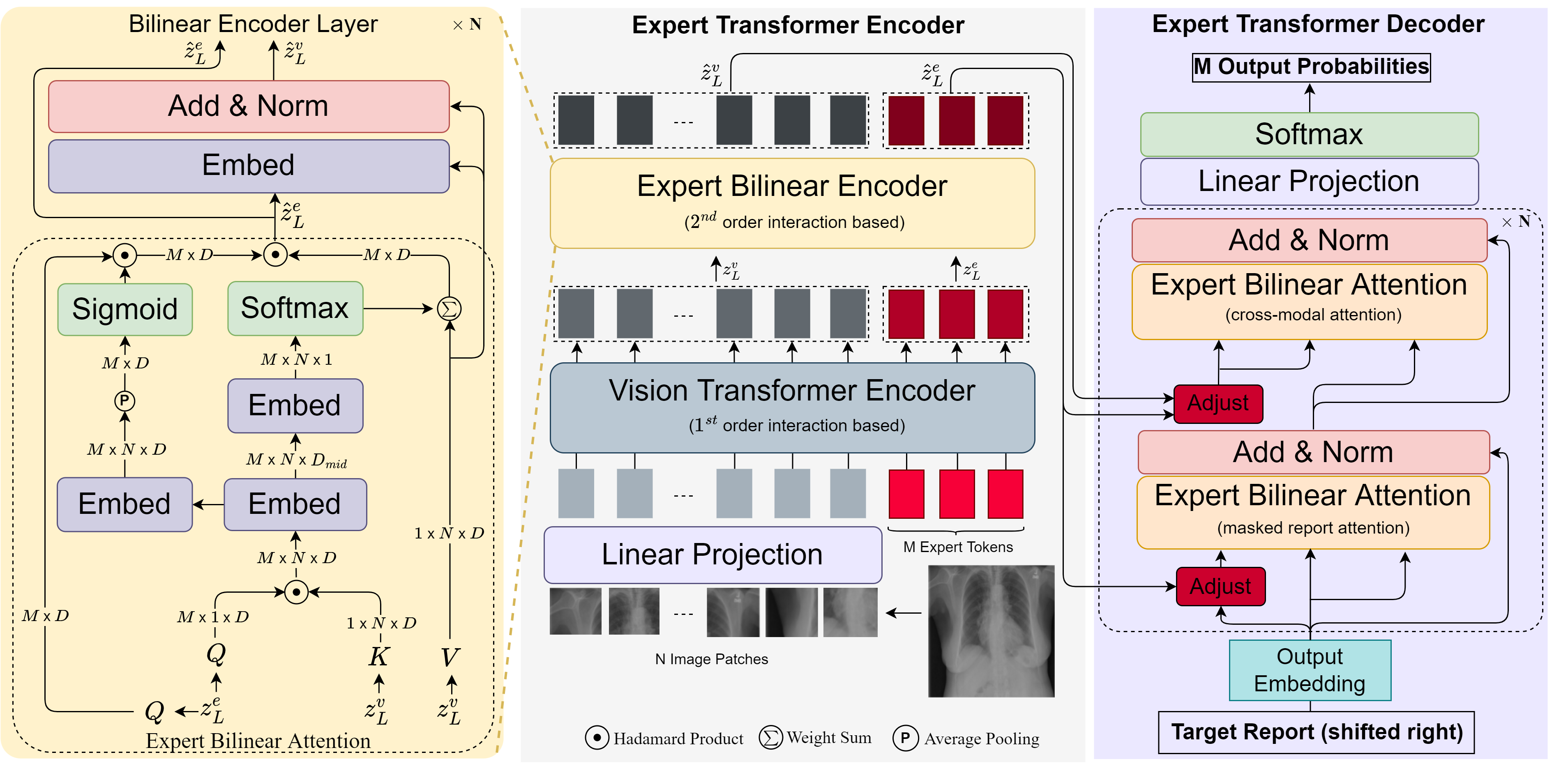}}
      \caption{An overview of our proposed METransformer, which includes an Expert Transformer Encoder and an Expert Transformer Decoder. The expert transformer encoder captures fine-grained visual patterns by exploring 1st and 2nd order interactions between input visual tokens and M expert tokens.
      The expert transformer decoder produces M diverse candidate reports guided by each expert token through an Adjust block. Noted that the ``Embed" layer includes a linear projection followed by a ReLU activation function.}
    \label{fig:framework}
    \vspace{-1em}
\end{figure*}

As shown in Figure.\ref{fig:framework}, our encoder comprises a vision transformer (ViT)~\cite{dosovitskiy2020vit} encoder and an expert bilinear transformer encoder. The ViT encoder takes both the expert and the visual tokens as the input and computes the linear attention between every two tokens. The encoded tokens are further sent into the expert bilinear transformer encoder where high-order interactions are computed between the expert tokens and the visual tokens.  The enhanced expert tokens are then used to regulate the embeddings of visual and word tokens by an adjust block and sent into the expert decoder to produce expert-specific reports in parallel, and then an expert voting strategy is used to generate the final report.


\label{sec:method}
\subsection{Multi-expert ViT Encoder}
Our ViT encoder adopts vision transformer~\cite{dosovitskiy2020vit}. In addition to the common input of visual patches/tokens embedding and position embedding, we further introduce expert tokens embedding and segment embedding. 

\textbf{Visual patches Embedding.} Given an image $\mathbf{x} \in \mathbb{R}^{H \times W \times C}$, followed ~\cite{dosovitskiy2020vit}, we reshape $\mathbf{x}$ into a sequence of flattened 2D patches $\mathbf{x}_p \in \mathbb{R}^{N \times (P^2 \dot C)}$, where $(H, W)$ is the resolution of the original image, $C$ is the number of channels, $(P, P)$ is the resolution of each image patch, and $N = HW/P^2$ is the resulting number of patches. We consider $\mathbf{x}_p$ as visual tokens and $N$ is the input sequence length.

\textbf{Expert tokens Embedding.} In addition to $N$ visual tokens, we post-pend $M$ learnable embeddings $\mathbf{x}_e \in \mathbb{R}^{M \times D}$ which have the same dimension as the visual token embeddings and are called expert tokens. We introduce an Orthogonal Loss as in Eqn.~\ref{loss:orl} to enforce orthogonality among the expert token embeddings, encouraging different expert tokens to attend different image regions.

\textbf{Segment Embedding.} We introduce two types of the segment, ``[0]" and ``[1]", to separate the input tokens from different sources~\cite{su2019vlbert}, i.e., ``[0]" for visual tokens and ``[1]" for expert tokens. A learned segment embedding $\mathbf{E}_{seg}$ is added to each input token to indicate its segment type.

\textbf{Position Embedding.} A standard 1D learnable position embedding $\mathbf{E}_{pos}$ is added to each input token to indicate its order in the input sequence. 

\textbf{Model Structure.} The expert ViT encoder adopts the same structure as the standard Transformer~\cite{vaswani2017attentionisallyouneed}, which consists of alternating layers of multi-headed self-attention (MSA) and Multi-Layer Perceptron blocks (MLP). Layernorm(LN)~\cite{ba2016layer} is applied before every block and residual connections are applied after every block. Mathematically:
\begin{equation}
\begin{aligned}
    \vspace{-2mm}
    \mathbf{z}_0 &= [\mathbf{x}_p^1\mathbf{E}; \mathbf{x}_p^2\mathbf{E}; \dots; \mathbf{x}_p^N\mathbf{E};\mathbf{x}_e^1; \mathbf{x}_e^2; \dots; \mathbf{x}_e^M]  \\
                &\quad+ \mathbf{E}_{pos} + \mathbf{E}_{seg}, \\
    \hat{\mathbf{z}_l} &= \mathrm{MSA}(\mathrm{LN}(\mathbf{z}_{l-1})) + \mathbf{z}_{l-1}, \\
    \mathbf{z}_l &= \mathrm{MLP}(\mathrm{LN}(\hat{\mathbf{z}_l})) + \hat{\mathbf{z}_l}, \\
\end{aligned}
\end{equation}
where $\mathbf{E} \in \mathbb{R}^{(P^2 C) \times D}$ is a learnable matrix parameter to map visual tokens to a fixed dimension D. The subscript $l = 1 \dots L$ and $L$ is the total number of the transformer layers. $\mathbf{E}_{pos} \in \mathbb{R}^{(N+M) \times D}$ is the position embeddings and $\mathbf{E}_{seg} \in \mathbb{R}^{(N+M) \times D}$ is the segment embeddings.

\subsection{Multi-expert Bilinear Attention Encoder}
The output of our expert ViT encoder, $\mathbf{z}_L \in \mathbb{R}^{(N+M) \times D}$, combines the visual token embeddings $\mathbf{z}_L^v = \mathbf{z}_L\left[:\mathrm{N} \right]$ and the expert token embeddings $\mathbf{z}_L^e = \mathbf{z}_L\left[\mathrm{N}:(\mathrm{N}+\mathrm{M})\right]$. They have attended to each other using multi-head self-attention, which is linear attention. Considering the fine-grained nature of medical images and the effectiveness of high-order attention in fine-grained recognition~\cite{lin2015bilinear}, we further enhance both embeddings by expert bilinear attention.

\textbf{Expert Bilinear Attention.} We followed ~\cite{pan2020xlinear} to design our expert bilinear attention (EBA) module. As shown in Figure.~\ref{fig:framework} Left, we take the enhanced expert token embeddings $\mathbf{z}_L^e$ as query $\mathbf{Q} \in \mathbf{R}^{M \times 1 \times D_q}$, the enhanced visual tokens embeddings $\mathbf{z}_L^v$ as key $\mathbf{K} \in \mathbf{R}^{1 \times N \times D_k}$ and value $\mathbf{V} \in \mathbf{R}^{1 \times N \times D_v}$, a low-rank bilinear pooling~\cite{lowrank} if first performed to obtain the joint bilinear query-key $\mathbf{B}_k$ and query-value $\mathbf{B}_v$ by $\mathbf{B}_k = \sigma(\mathbf{W}_k \mathbf{K}) \odot \sigma(\mathbf{W}_q^k\mathbf{Q})$ and $\mathbf{B}_v = \sigma(\mathbf{W}_v \mathbf{V}) \odot \sigma(\mathbf{W}_q^v\mathbf{Q})$, where $\mathbf{W}_k \in \mathbf{R}^{D_B \times D_k}$, $\mathbf{W}_v \in \mathbf{R}^{D_B \times D_v}$, $\mathbf{W}_q^k \in \mathbf{R}^{D_B \times D_q}$ and $\mathbf{W}_q^v \in \mathbf{R}^{D_B \times D_v}$ are learnable parameters, resulting $\mathbf{B}_k \in \mathbb{R}^{M \times N \times D_B}$, $\mathbf{B}_v \in \mathbb{R}^{M \times N \times D_B}$. $\sigma$ denotes ReLU unit, and $\odot$ represents Hadamard Product. Afterward, we compute attention for input tokens both spatial and channel-wisely. 1) Spatial-wise attention. We use a linear layer to project ${\mathbf{B}_k}$ into an intermediate representation ${\mathbf{B}_{mid}} = \sigma(\mathbf{W}_B^k \mathbf{B}_k)$, where $\mathbf{W}_B^k \in \mathbb{R}^{D_B \times D_{mid}}$. Then another linear layer is applied to mapping $\mathbf{B}_{mid} \in \mathbb{R}^{M \times N \times D_{mid}}$ from the dimension $\mathbf{D}_{mid}$ to 1 and further followed a softmax function to obtain the spatial-wise attention weight $\alpha_s \in \mathbb{R}^{M \times N \times 1}$. 2) Channel-wise attention. A squeeze-excitation operation~\cite{hu2018squeeze} to ${\mathbf{B}_{mid}}$ is performed to obtain channel-wise attention $\beta_c = \mathrm{sigmoid}(\mathbf{W}_c \bar{\mathbf{B}_{mid}})$, where $\mathbf{W}_c \in \mathbf{R}^{\mathbf{D}_{mid} \times \mathbf{D}_B}$ is learnable parameters and $\bar{\mathbf{B}_{mid}} \in \mathbf{R}^{M \times \mathbf{D}_{mid}}$ is the average pooling of ${\mathbf{B}_{mid}}$. 
The first layer of EBA is formulated as,
\begin{equation}
    \vspace{-2mm}
    \hat{\mathbf{z}}_{L}^{e(1)} = \mathrm{EBA}(\hat{\mathbf{z}}_{L}^{e}, \hat{\mathbf{z}}_{L}^{v}) = \beta_c \odot  \alpha_s \mathbf{B}_v.
\end{equation}

\textbf{Bilinear Encoder Layer.} Besides EBA, we also use the ``Add \& Norm" layer with the residual connection as in a standard transformer in our bilinear encoder layer. The $n$-th layer bilinear encoder is expressed as,
\begin{equation}
\begin{aligned}
\vspace{-2mm}
    \hat{\mathbf{z}}_{L}^{e(n)} &= \mathrm{EBA}(\hat{\mathbf{z}}_{L}^{e(n-1)}, \hat{\mathbf{z}}_{L}^{v(n-1)}) \\
    \hat{\mathbf{z}}_{L}^{v(n)} &= \mathrm{LN}(\mathbf{W}_e^n[\hat{\mathbf{z}}_{L}^{e(n-1)};\hat{\mathbf{z}}_{L}^{v(n-1)}] + \hat{\mathbf{z}}_{L}^{v(n-1)})
\end{aligned}
\end{equation}
 where $n=1 \dots N$ and $N$ is the number of bilinear encoder layer, $\mathrm{LN}(\cdot)$ denotes layer normalization~\cite{ba2016layer}, $\mathbf{W}_e^n \in \mathbb{R}^{(D_q + D_v) \times D_v}$ is learnable parameters and $[;]$ denotes concatenation. We set $\hat{\mathbf{z}}_{L}^{e(0)}=\mathbf{z}_L^e$ and $\hat{\mathbf{z}}_{L}^{v(0)}=\mathbf{z}_L^v$ when n=1.
\subsection{Multi-expert Bilinear Attention Decoder}
The output of the last expert bilinear encoder, expert token embeddings $\hat{\mathbf{z}}_L^{e(N)} \in \mathbb{R}^{N \times D_B}$ and visual tokens embeddings $\hat{\mathbf{z}}_L^{v(N)} \in \mathbb{R}^{M \times D_B}$, are sent to the decoder for report generation. The bilinear decoder layer comprises an $\mathrm{EBA}_{mask}$ layer to compute attention of the shifted-right reports and an $\mathrm{EBA}_{cross}$ to compute the cross-modal attention and an adjust block to regulate the word and visual tokens embeddings by expert token embeddings. For convenience, we denote $\hat{\mathbf{z}}_L^{e(N)}$ and $\hat{\mathbf{z}}_L^{v(N)}$ as $\mathbf{f}_e$ and $\mathbf{f}_v$.

\textbf{Adjust block.} To incorporate the expert tokens into the report generation process, we propose an expert adjustment block that allows each expert token embedding to influence the embedding of the words and visual tokens, thereby generating the report associated with that expert token. Since our expert tokens are trained orthogonally, we can generate discrepant reports by the different expert tokens. To regulate visual tokens embeddings $\mathbf{f}_v$ by expert token embeddings $\mathbf{f}_e$, the adjust block is calculated as follows,
\begin{equation}
    \hat{\mathbf{f}}_v = \mathrm{F_{adjust}}(\mathbf{f}_{e}, \mathbf{f}_v) = \sigma(\mathbf{W}_e \mathbf{f}_e) \odot \sigma(\mathbf{W}_v \mathbf{f}_v)
\end{equation}
where $\mathbf{W}_{e}$ and $\mathbf{W}_v$ are learnable parameters. $\sigma$ denotes ReLU unit, and $\odot$ represents Hadamard Product.

\textbf{Bilinear Decoder Layer.} We first perform a masked EBA to word embeddings $\mathbf{E}_r$ and then followed another EBA block to compute the cross attention of word embeddings and visual tokens embeddings. We also employ residual connections around each EBA block similar to the encoder, followed by layer normalization. Mathematically, the $i$-th layer bilinear decoder can be expressed as,
\begin{equation}
\begin{aligned}
    \mathbf{E}_{mid}^{(i)} &= \mathrm{LN}(\mathrm{EBA}_{mask}(\hat{\mathbf{E}}_r^{(i-1)}, \mathbf{E}_r^{(i-1)}) + \mathbf{E}_r^{(i-1)}) \\
    \mathbf{E}_c^{(i)} &= \mathrm{LN}(\mathrm{EBA}_{cross}(\mathbf{E}_{mid}^{(i)}, \hat{\mathbf{f}}_v) + \mathbf{E}_{mid}^{(i)}) \\
    \mathbf{E}_r^{(i)} &= \mathrm{LN}(\mathbf{W}_d^i[\mathbf{E}_r^{(i-1)};\mathbf{E}_c^{(i)}] + \mathbf{E}_r^{(i-1)}))
\end{aligned}
\end{equation}
where $i=1 \dots I$ and $I$ represents the total number of decoder layer. $\hat{\mathbf{E}}_r^{(i-1)} = \mathrm{F_{adjust}}(\mathbf{f}_e, \mathbf{E}_r^{(i-1)})$ and $\hat{\mathbf{f}}_v = \mathrm{F_{adjust}}(\mathbf{f}_e, \mathbf{f}_v)$. $\mathbf{W}_d^i \in \mathbb{R}^{(D_r + D_r) \times D_r}$ is learnable parameters and $[;]$ denotes concatenation. Specifically, $E_r^{(0)} \in \mathbb{R}^{T \times D_r}$ is the original word embeddings where T is the total number of words in the reports and $D_r = D_B$ is the dimension of word embedding, and we extend it to M replicates corresponding to M expert tokens to compute parallelly. The final output of the decoder is $E_c^I \in \mathbb{R}^{M \times T \times D_r}$, which will be further used to predict word probabilities by a linear projection and the softmax activation function.
\subsection{Objective Function}
\textbf{Orthogonal Loss.} To encourage orthogonality among expert token embeddings, we introduce an orthogonal loss term to the output of expert bilinear encoder $\hat{\mathbf{z}}_L^{e}$,
\begin{equation}
    \mathcal{L}_{OrL}(\hat{\mathbf{z}}_L^{e}) = \frac{1}{M} \left \| \ell(\hat{\mathbf{z}}_L^{e})^{\top}\ell(\hat{\mathbf{z}}_L^{e}) - I \right \|^2
\label{loss:orl}
\end{equation}
where $\ell(\cdot)$ denotes $L_2$ normalization and $I$ represents the identity matrix with dimension M.

\textbf{Report Generation Loss.} We train our model parameters $\theta$ by minimizing the negative log-likelihood of $\mathbf{P}(t)$ given the image features: 
\begin{equation}
    {\mathcal L}_{CE} = -\frac{1}{M}\sum_{m=1}^{M}\sum_{i=1}^{T}log P_{\theta}({\mathbf t}_i^{(m)}|{\mathbf I}, {\mathbf t}_{i-1}^{(m)}, \cdots, {\mathbf t}_1^{(m)})
\end{equation}
where $\mathbf{P}({\mathbf t}_i^{(m)}|{\mathbf I}, {\mathbf t}_{i-1}^{(m)}, \cdots, {\mathbf t}_1^{(m)})$ represents the probability predicted by the $m$-th expert tokens for the  $i$-th word $t_i$ based on the image $\mathbf{I}$ and the first $(i-1)$ words.

Our overall objective function is:
  $ {\mathcal L}_{all} = {\mathcal L}_{CE} + \lambda {\mathcal L}_{OrL}$.
The hyper-parameter $\lambda$ simply balances the two loss terms, and its value is given in Section~\ref{sec:experiments}. 
\subsection{Expert Voting strategy}
For the $M$ diagnostic reports $R = \left[r_1, r_2, \dots, r_M\right]$ produced by $M$ experts, we design a metric-based expert voting strategy to select the optimal one, where the ``metric" means the conventional natural language generation (NLG) metrics, such as BLEU-4 and CIDEr (we use CIDEr in our paper). The voting score $\mathbf{S}_i$ for $i$-th expert's report can be calculated by the following equation:
\begin{equation}
\label{eq:metric}
    \mathbf{S}_{i} = {\textstyle \sum_{j=1, j\ne i}^{M}}\mathrm{CIDEr}(\mathbf{r}_{i},\mathbf{r}_j)
\end{equation}
where $\mathrm{CIDEr(,)}$ denotes the function for computing CIDEr with $\mathbf{r}_i$ as candidate and $\mathbf{r}_j$ as reference. In this way, each expert's report can get a vote score, indicating the degree of consistency of the report with that of other experts.The diagnostic report with the highest score is the winner of the voting.
As demonstrated in Table.~\ref{tab:ablation_study}, our voting strategy is more effective than the commonly used ensemble/fusion methods~\cite{pan2020xlinear}. The possible reasons are that i) our method utilizes NLG metrics as reference scores, thereby the voted results are directly related to the final evaluation metrics; ii) since we ultimately select a single result with the highest score, we can produce a more coherent report than fusing multiple results at the word level.
\begin{table*}[t]
{\small
\begin{tabular*}{\hsize}{@{}@{\extracolsep{\fill}}l|l|ccccccc@{}}
\hline
Dataset   & Methods       & BLEU-1         & BLEU-2         & BLEU-3         & BLEU-4         & ROUGE  & METEOR         & CIDEr          \\ \hline
          & Show-Tell~\cite{vinyals2015showandtell}     & 0.243           & 0.130          & 0.108          & 0.078          & 0.307  & 0.157              &  0.197        \\
          & Att2in~\cite{rennie2017selfcritical}        & 0.248          & 0.134          & 0.116          & 0.091          & 0.309  & 0.162              & 0.215        \\
          & AdaAtt~\cite{2017Knowing}        & 0.284          & 0.207          & 0.150          & 0.126          & 0.311  & 0.165              & 0.268         \\
          & Transformer~\cite{vaswani2017attentionisallyouneed}   & 0.372          & 0.251          & 0.147          & 0.136          & 0.317  & 0.168              & 0.310         \\
          & M2transformer\cite{2020Meshed} & 0.402          & 0.284          & 0.168          & 0.143          & 0.328          & 0.170      & 0.332      \\
          & R2Gen$^{\dagger}$~\cite{chen2020generating}          & 0.470          & 0.304          & 0.219          & 0.165          & 0.371  & 0.187       & -                 \\
          & R2GenCMN$^{\dagger}$~\cite{chen2022cross}          & 0.475          & 0.309          & 0.222          & 0.170          & 0.375  & 0.191              & -          \\
IU-Xray   & MSAT~\cite{wang2022medical}         & 0.481           & 0.316          & 0.226          & 0.171          & 0.372  & 0.190          & 0.394              \\
          \cline{2-9} 
          & Ours(METransformer)          & \textbf{0.483}          & \textbf{0.322}          & \textbf{0.228}          & \textbf{0.172}          & \textbf{0.380}  & \textbf{0.192}               & \textbf{0.435} \\
          \cline{2-9} 
          & \multicolumn{8}{c}{Results below are not strictly comparable due to different data partition. For reference only.}  \\ \cline{2-9}
          & {\color[HTML]{9B9B9B} CoAtt$^{\dagger}$~\cite{2017Ontheautomatic}}         & {\color[HTML]{9B9B9B} 0.455}          & {\color[HTML]{9B9B9B} 0.288}          & {\color[HTML]{9B9B9B} 0.205}          & {\color[HTML]{9B9B9B} 0.154}          & {\color[HTML]{9B9B9B} 0.369}  & -              & {\color[HTML]{9B9B9B} 0.277}          \\
          & {\color[HTML]{9B9B9B} HGRG-Agent$^{\dagger}$ \cite{li2018hybrid}}    & {\color[HTML]{9B9B9B} 0.438}          & {\color[HTML]{9B9B9B} 0.298}          & {\color[HTML]{9B9B9B} 0.208}          & {\color[HTML]{9B9B9B} 0.151}          & {\color[HTML]{9B9B9B} 0.322}  & {\color[HTML]{9B9B9B} -}              & {\color[HTML]{9B9B9B} 0.343}           \\
          & {\color[HTML]{9B9B9B} KERP$^{\dagger}$ \cite{Knowledge-Driven}}          & {\color[HTML]{9B9B9B} 0.482}          & {\color[HTML]{9B9B9B} 0.325}          & {\color[HTML]{9B9B9B} 0.226}          & {\color[HTML]{9B9B9B} 0.162}          & {\color[HTML]{9B9B9B} 0.339}  & {\color[HTML]{9B9B9B} -}              & {\color[HTML]{9B9B9B} 0.280}         \\
          & {\color[HTML]{9B9B9B} PPKED$^{\dagger}$~\cite{CVPR201_PPKD}}          & {\color[HTML]{9B9B9B} 0.483}          & {\color[HTML]{9B9B9B} 0.315}          & {\color[HTML]{9B9B9B} 0.224}          & {\color[HTML]{9B9B9B} 0.168}          & {\color[HTML]{9B9B9B} 0.376}  & {\color[HTML]{9B9B9B} 0.187}              & {\color[HTML]{9B9B9B} 0.351}          \\
          & {\color[HTML]{9B9B9B} GSK$^{\dagger}$~\cite{2021Knowledge}}         & {\color[HTML]{9B9B9B} 0.496}           & {\color[HTML]{9B9B9B} 0.327}          & {\color[HTML]{9B9B9B} 0.238}          & {\color[HTML]{9B9B9B} 0.178}          & {\color[HTML]{9B9B9B} 0.381}  & {\color[HTML]{9B9B9B} -}         & {\color[HTML]{9B9B9B} 0.382}              \\ \hline
          \hline
          & Show-Tell~\cite{vinyals2015showandtell}     & 0.308         & 0.190         & 0.125         & 0.088         & 0.256 & 0.122         & 0.096         \\
          & Att2in~\cite{rennie2017selfcritical}        & 0.314         & 0.198         & 0.133         & 0.095         & 0.264  & 0.122         & 0.106         \\
          & AdaAtt~\cite{2017Knowing}        & 0.314         & 0.198         & 0.132         & 0.094         & 0.267 & 0.128          & 0.131         \\
          & Transformer~\cite{vaswani2017attentionisallyouneed}   & 0.316              & 0.199              & 0.140              & 0.092              & 0.267      & 0.129            & 0.134              \\
          & M2Transformer~\cite{2020Meshed} & 0.332              & 0.210              & 0.142              & 0.101              & 0.264      & 0.134              & 0.142              \\
          & R2Gen$^{\dagger}$~\cite{chen2020generating}         & 0.353          & 0.218          & 0.145          & 0.103          & 0.277  & 0.142          & -          \\
MIMIC-CXR & R2GenCMN$^{\dagger}$~\cite{chen2022cross}         & 0.353          & 0.218          & 0.148          & 0.106          & 0.278  & 0.142          & -          \\
          & PPKED$^{\dagger}$~\cite{CVPR201_PPKD}         & 0.36           & 0.224          & 0.149          & 0.106          & 0.284  & 0.149          & 0.237              \\
          & GSK$^{\dagger}$~\cite{2021Knowledge}         & 0.363           & 0.228          & 0.156          & 0.115          & 0.284  & -          & 0.203              \\
          & MSAT$^{\dagger}$~\cite{wang2022medical}         & 0.373           & 0.235          & 0.162          & 0.120          & 0.282  & 0.143          & 0.299              \\
          \cline{2-9} 
          & Ours(METransformer)          & \textbf{0.386} & \textbf{0.250} & \textbf{0.169} & \textbf{0.124} & \textbf{0.291}  & \textbf{0.152} & \textbf{0.362} \\ \hline
\end{tabular*}
}
\vspace{-2mm}
\caption{Comparison on IU-Xray (upper part) and MIMIC-CXR datasets (lower part). $\dagger$ indicates the results are quoted from the published literature: the results of CoAtt~\cite{2017Ontheautomatic} and HGRG-Agent~\cite{li2018hybrid} on IU-Xray are quoted from ~\cite{Knowledge-Driven} while the other results are quoted from their respective papers.  For the methods without $\dagger$, their results are obtained by re-running the publicly released codebase~\cite{li2021codebase} on these two datasets using the same training-test partition as our method. 
}\label{Table:ComparisonWithSOTA}
\vspace{-4mm}
\end{table*}

\section{Experiments}
\label{sec:experiments}

\subsection{Datasets}
In this experiment, two datasets are used for the performance evaluation, i.e., a widely-used benchmark IU-Xray~\cite{2015iu-xray} and the currently largest dataset MIMIC-CXR~\cite{2019MIMIC} for medical report generation.

\textbf{IU-Xray} Indiana University Chest X-ray Collection (IU-Xray) \cite{2015iu-xray} is the most widely used publicly accessible dataset in medical report generation tasks. It contains 3,955 fully de-identified radiology reports, each of which is associated with frontal and/or lateral chest X-ray images, and 7,470 chest X-ray images in total. Each report is comprised of several sections: Impression, Findings, Indication, etc. In this work, we adopt the same data set partitioning as~\cite{chen2020generating} for a fair comparison, with a train/test/val set by 7:1:2 of the entire dataset.  All evaluations are done on the test set.

\textbf{MIMIC-CXR} The recently released MIMIC-CXR~\cite{2019MIMIC} is the largest public dataset containing both chest radiographs and free-text reports. In total, it consists of 377110 chest x-ray images and 227835 reports from 64588 patients of the Beth Israel Deaconess Medical Center examined between 2011 and 2016. In our experiment, we adopt MIMIC-CXR's official split following the work~\cite{chen2020generating} for a fair comparison, resulting in a total of 222758 samples for training, and 1808 and 3269 samples for validation and test.

\subsection{Experimental Settings}
\textbf{Evaluation Metrics}~~Following the standard evaluation protocol\footnote{https://github.com/tylin/coco-caption}, we utilize the most widely used BLEU-4~\cite{Kishore2002bleu}, METEOR~\cite{banerjee-lavie-2005-meteor}, ROUGE-L~\cite{lin-2004-rouge}, and CIDEr~\cite{vedantam2015cider} as the metrics to evaluate the quality of the generated diagnostic reports. 
To measure the accuracy of descriptions for clinical abnormalities, we follow~\cite{chen2020generating, chen2022cross, 2021Auto} and further report clinical efficacy metrics. For this purpose, the CheXpert~\cite{irvin2019chexpert} is applied to labeling the generated reports and the results are compared with ground truths in 14 different categories related to thoracic diseases and support devices. We use precision, recall, and F1 to evaluate model performance for clinical efficacy metrics.

\begin{table}[ht]
\centering
\setlength{\tabcolsep}{2mm}{
\begin{tabular}{c|ccc}
\hline
Methods     & Precision & Recall & F1    \\ \hline
Show-Tell~\cite{vinyals2015showandtell}   & 0.249     & 0.203  & 0.204 \\
Att2in~\cite{xu2016showattandtell}      & 0.268     & 0.186  & 0.181 \\
AdaAtt~\cite{2017Knowing}      & 0.322     & 0.239  & 0.249 \\
Transformer~\cite{vaswani2017attentionisallyouneed} & 0.331     & 0.224  & 0.228 \\
R2Gen~\cite{chen2020generating}       & 0.333     & 0.273  & 0.276 \\
R2GenCMN~\cite{chen2022cross}    & 0.334     & 0.275  & 0.278 \\ \hline
Ours(METransformer) & \textbf{0.364}     & \textbf{0.309}  & \textbf{0.311} \\ \hline
\end{tabular}}
\caption{Comparison of clinical efficacy metrics on the test set of the MIMIC-CXR dataset for measuring the accuracy of the description of clinical abnormalities.}~\label{tab:ce_metrics}
\vspace{-8mm}
\end{table}
\begin{table*}[ht]
\centering
\scalebox{0.85}{
\begin{tabular}{llcccccccccc}
\hline
\multicolumn{1}{c}{\multirow{2}{*}{\#}} & \multicolumn{1}{c}{\multirow{2}{*}{Models}} & \multicolumn{5}{c}{IU-Xray}                      & \multicolumn{5}{c}{MIMIC-CXR}                    \\ \cline{3-12} 
\multicolumn{1}{c}{}                    & \multicolumn{1}{c}{}                        & BLEU\_4 & ROUGE & METEOR & CIDEr & AVG. $\Delta$ & BLEU\_4 & ROUGE & METEOR & CIDEr & AVG. $\Delta$ \\ \hline
1                                       & \multicolumn{1}{l|}{BASELINE}               & 0.161   & 0.357 & 0.183  & 0.337 & -             & 0.109   & 0.277 & 0.143  & 0.275 & -             \\
2                                       & \multicolumn{1}{l|}{+BE}                    & 0.163   & 0.360 & 0.185  & 0.346 & +1.5\%        & 0.111   & 0.279 & 0.144  & 0.287 & +1.9\%        \\
3                                       & \multicolumn{1}{l|}{+BE+ETs}                & 0.168   & 0.372 & 0.188  & 0.402 & +7.6\%        & 0.117   & 0.284 & 0.147  & 0.334 & +8.5\%        \\
4                                       & \multicolumn{1}{l|}{+BE+ETs+OrL}            & 0.170   & 0.377 & 0.190  & 0.426 & +10.4\%       & 0.121   & 0.287 & 0.150  & 0.352 & +11.9\%       \\
5                                       & \multicolumn{1}{l|}{+BE+ETs+OrL+EV}         & 0.172   & 0.380 & 0.192  & 0.435 & +11.8\%       & 0.124   & 0.291 & 0.152  & 0.362 & +14.2\%       \\ \hline
\end{tabular}}
{\vspace{-2mm}}
\caption{Ablation studies. The ``BASELINE" uses ViT encoder and bilinear-attention decoder with the traditional cross entropy-loss. Here ``BE", ``ETs", ``OrL", ``EV" stand for "Bilinear-attention Encoder", "Expert Tokens", "Orthogonal Loss", and "Expert Voting". 
The average improvement over all NLG metrics compared to baseline
is also presented in the ``AVG. $\Delta$" column.}~\label{tab:ablation_study}
\vspace{-4mm}
\end{table*}
\begin{table*}[ht]
\centering
\setlength{\tabcolsep}{2mm}{
\begin{tabular}{l|c|c|cccc}
\hline
\multicolumn{1}{c|}{Models} & Params  & Training Time & BLEU\_4 & ROUGE & METEOR & CIDEr \\ \hline
METransformer(num\_expert=1)      & 152M & 0.48h       & 0.163   & 0.362 & 0.183  & 0.346 \\ \hline
Stochastic Model Averaging      & 152M x 7=1064M & 0.48h x 7=3.36h       & 0.168   & 0.373 & 0.187  & 0.389 \\
Random Initialization      & 152M x 7=1064M & 0.48h x 7=3.36h       & 0.170   & 0.376 & 0.189  & 0.408 \\
Multiple Decoder           & 483.3M    & 1.85h       & 0.166   & 0.368 & 0.186  & 0.378 \\
METransformer(num\_expert=7)        & 152.007M    & 0.55h       & \textbf{0.172}   & \textbf{0.380} & \textbf{0.192}  & \textbf{0.435} \\ \hline
\end{tabular}}
\vspace{-2mm}
\caption{Comparison with the ensemble models on the IU-Xray dataset. 
}
\vspace{-4mm}
\label{tab:ensemble}
\end{table*}
\textbf{Implementation Details}~~For IU-Xray, we use image pairs for report generation as~\cite{chen2020generating}. 
For both datasets, we use the "bert-base-uncased" model's tokenizer in huggingface transformer~\cite{wolf-etal-2020-transformers} to tokenize all words in the reports. We utilize a pre-trained vision transformer with a patch size of 32 to initialize our Expert ViT encoder. The number of layers of the expert bilinear encoder and decoder is set as (2, 4) for MIMIC-CXR and (2,2) for IU-Xray to reduce the potential overfitting on IU-Xray due to its relatively small data size.  The dimensions of the bilinear query-key representation and the transformed bilinear feature ($D_B$ and $D_{mid}$) in the expert bilinear attention block is set as 768 and 384, respectively. The hyper-parameter $\lambda$ is set to 2. Our model is trained using Radam optimizer~\cite{liu2019radam} with a mini-batch size of 16. The learning rate is set to be 0.0001 and the model is trained for a total of 20 epochs. We implement our model using
Pytorch~\cite{paszke2019pytorch} and Pytorch-lightning library~\footnote{https://github.com/Lightning-AI/lightning} with two NVIDIA GeForce RTX 3090 GPU cards.
\subsection{Main Results}
Two types of metrics are used in our evaluation: the conventional natural language generation (NLG) metrics and the clinical efficacy (CE) metrics. The results are reported in Table~\ref{Table:ComparisonWithSOTA} and Table.~\ref{tab:ce_metrics}~\footnote{It is noted that clinical efficacy metrics only apply to MIMIC-CXR because the labeling schema of CheXpert is designed for MIMIC-CXR.}, respectively.

Specifically, we compare METransformer with 5 state-of-the-art (SOTA) image captioning methods, including Show-tell~\cite{vinyals2015showandtell}, AdaAtt~\cite{2017Knowing}, Att2in~\cite{topdown},   Transformer~\cite{devlin2019bert} and M2transformer~\cite{2020Meshed}. For these methods, we use a publicly released codebase~\cite{li2021codebase}  and re-run them on both datasets with the same experimental setting as ours. Moreover, eight SOTA medical report generation models are involved in the comparison, including CoAtt~\cite{2017Ontheautomatic}, HGRN-Agent~\cite{li2018hybrid}, R2Gen~\cite{chen2020generating}, R2GenCMN~\cite{chen2022cross}, PPKED~\cite{CVPR201_PPKD}, GSK~\cite{2021Knowledge} and MSAT~\cite{wang2022medical}. It is noteworthy that except R2Gen~\cite{chen2020generating} and R2GenCMN~\cite{chen2022cross}, these methods do not have their source code released. CoAtt~\cite{2017Ontheautomatic} and HGRN-Agent~\cite{li2018hybrid} only report results on IU-Xray in their original paper. For the others, we cite the results from their respective papers. Please note that since IU-Xray does not provide an official training-test partition, the cited performances of these methods on IU-Xray (except R2Gen~\cite{chen2020generating} and R2GenCMN~\cite{chen2022cross} that use the same partition as ours) are not strictly comparable,  and they are provided here only for reference. Differently, on MIMIC-CXR, since all these models follow the MIMIC-CXR official training-test partition, their cited performances are comparable.

As shown in Table~\ref{Table:ComparisonWithSOTA}, on both datasets, our METransformer consistently outperforms those ``single-expert" based models, including attention-based baselines (Att2in~\cite{xu2016showattandtell}, AdaAtt~\cite{2017Knowing}), memory-augmented methods (R2Gen~\cite{chen2020generating}, R2GenCMN~\cite{chen2022cross}) and models introducing external information (PPKED~\cite{CVPR201_PPKD}, MSAT~\cite{wang2022medical}). On MIMIC-CXR, METransformer is the best performer across all metrics. Especially, our CIDEr score is up to 0.362, which is to date the best performance and makes a significant improvement over the second best score of 0.299 of MSAT~\cite{wang2022medical}. These improvements demonstrate the advantage of our framework which is conceptually "multiple expert co-diagnosis". Ours METransformer also surpasses these methods on IU-Xray over most metrics, while it is again worth mentioning that the cited performances on IU-Xray may not be strictly comparable as obscure training-test partitions were used by different methods.
\subsection{Ablation Study}
\noindent\textbf{Contribution of each component.} We conduct an ablation study to single out the contributions of each model component, as presented in Table~\ref{tab:ablation_study}. We build a baseline by using ViT transformer encoder and bilinear-attention decoder to verify the performance improvements brought by multiple expert tokens, orthogonal loss, and our metric-based expert voting strategy. In Table~\ref{tab:ablation_study}, there are four components: ``BE", ``ETs", ``OrL", and ``EV", representing Bilinear-attention Encoder, Expert Tokens, Orthogonal Loss, and Expert Voting, respectively. The symbol ``+" indicates the inclusion of the following component based on the ``BASELINE" model. It can be observed that each component of METransformer has a positive effect on performance.
By comparing \#3 and \#1 in Table.~\ref{tab:ablation_study}, it can be found that extending the expert tokens and enhancing them with bilinear-attention encoder can increase the overall performance by 7.6\% on IU-Xray and 8.5\% on MIMIC-CXR. Training with the orthogonal loss on expert tokens can further enhance model performance (by comparing \#4 and \#3). 
For experiments \#3 and \#4 where multiple expert tokens are used, the final prediction result is obtained by averaging the probability of words predicted by the multiple experts. 
Comparing \#5 and \#4 shows that our voting strategy is more effective than the averaging method above.

\begin{figure}[t]
    \centering
    \centerline{\includegraphics[width=0.9\linewidth]{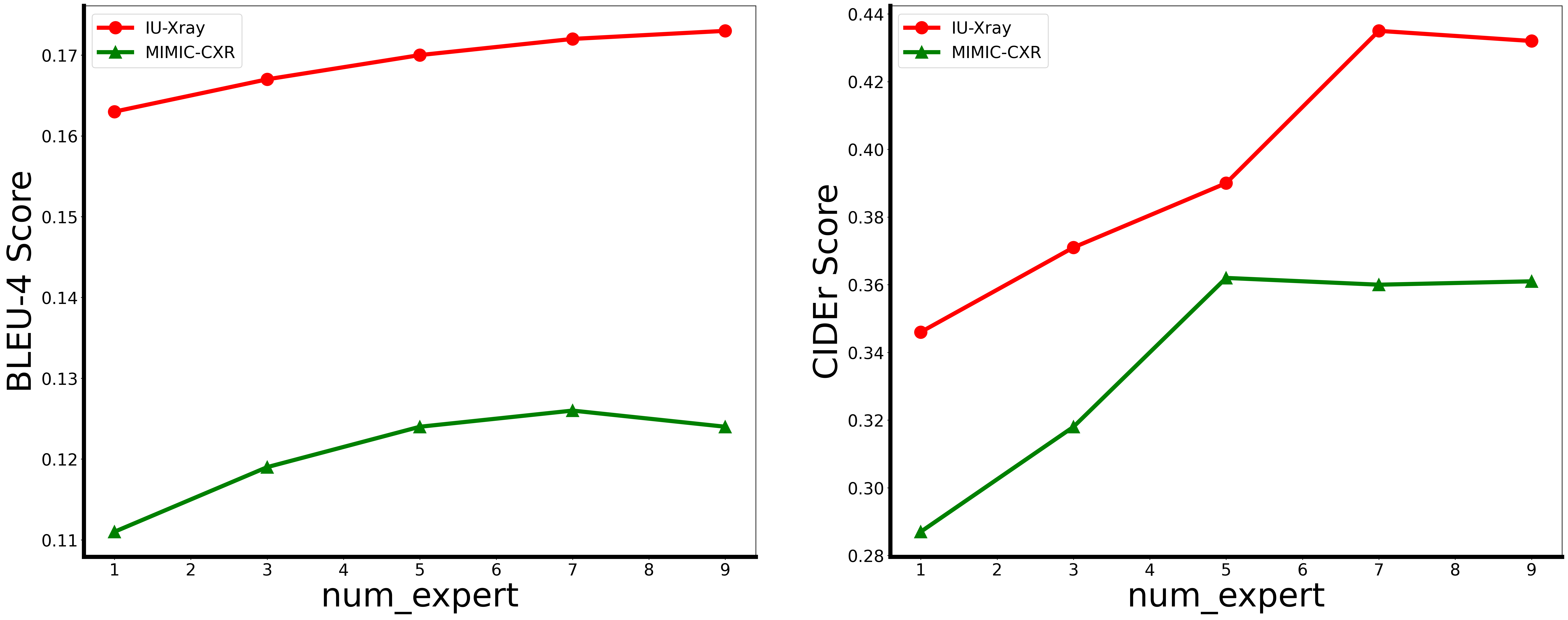}}
    \vspace{-2mm}
      \caption{Bleu\_4 and CIDEr scores by using different numbers of expert tokens on IU-Xray and MIMIC-CXR dataset.}
    \label{fig:num_tokens}
    \vspace{-6mm}
\end{figure}

\begin{figure*}[t]
\centering
\centerline{\includegraphics[width=0.9\linewidth]{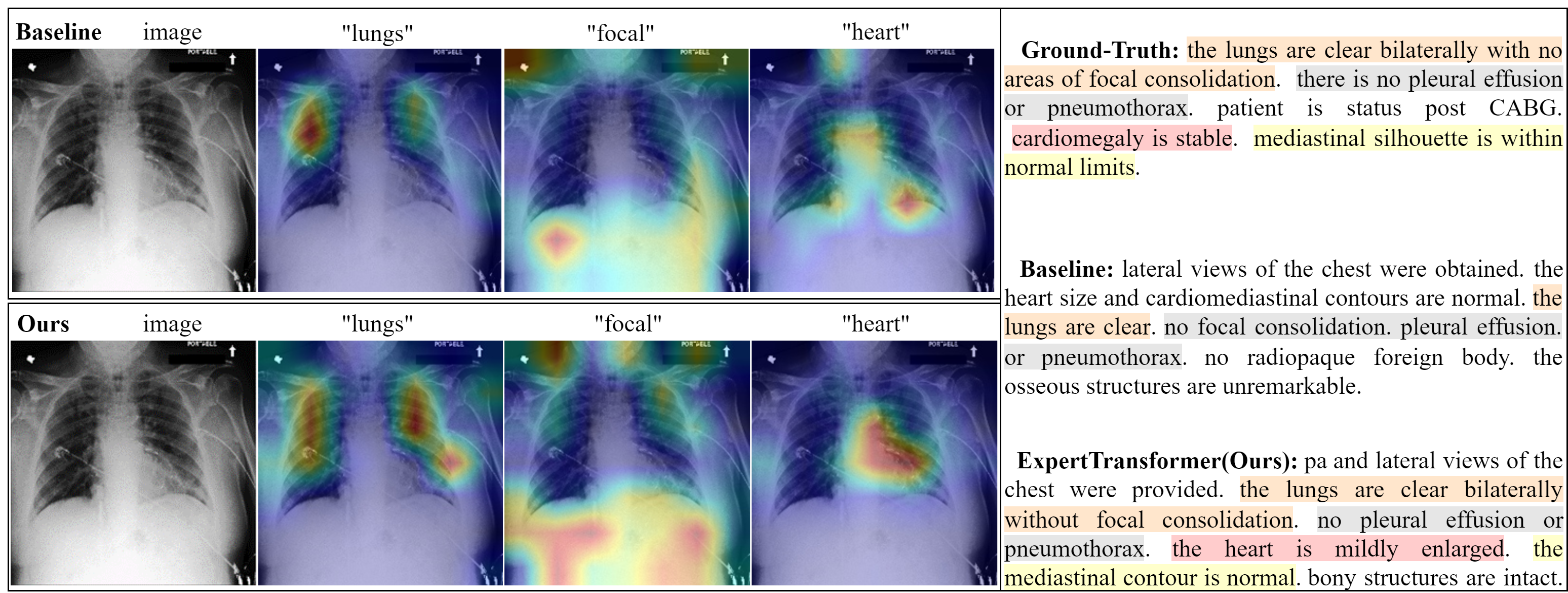}}
\caption{An example of the generated reports and their attention-mapping visualization of three key medical terms from BASELINE and ours METransformer. For better illustration, different colors in the generated reports highlight different medical information.}
\label{fig:vis_samples}
\vspace{-4mm}
\end{figure*}

\begin{figure}[t]
    \centering
    \centerline{\includegraphics[width=0.85\linewidth]{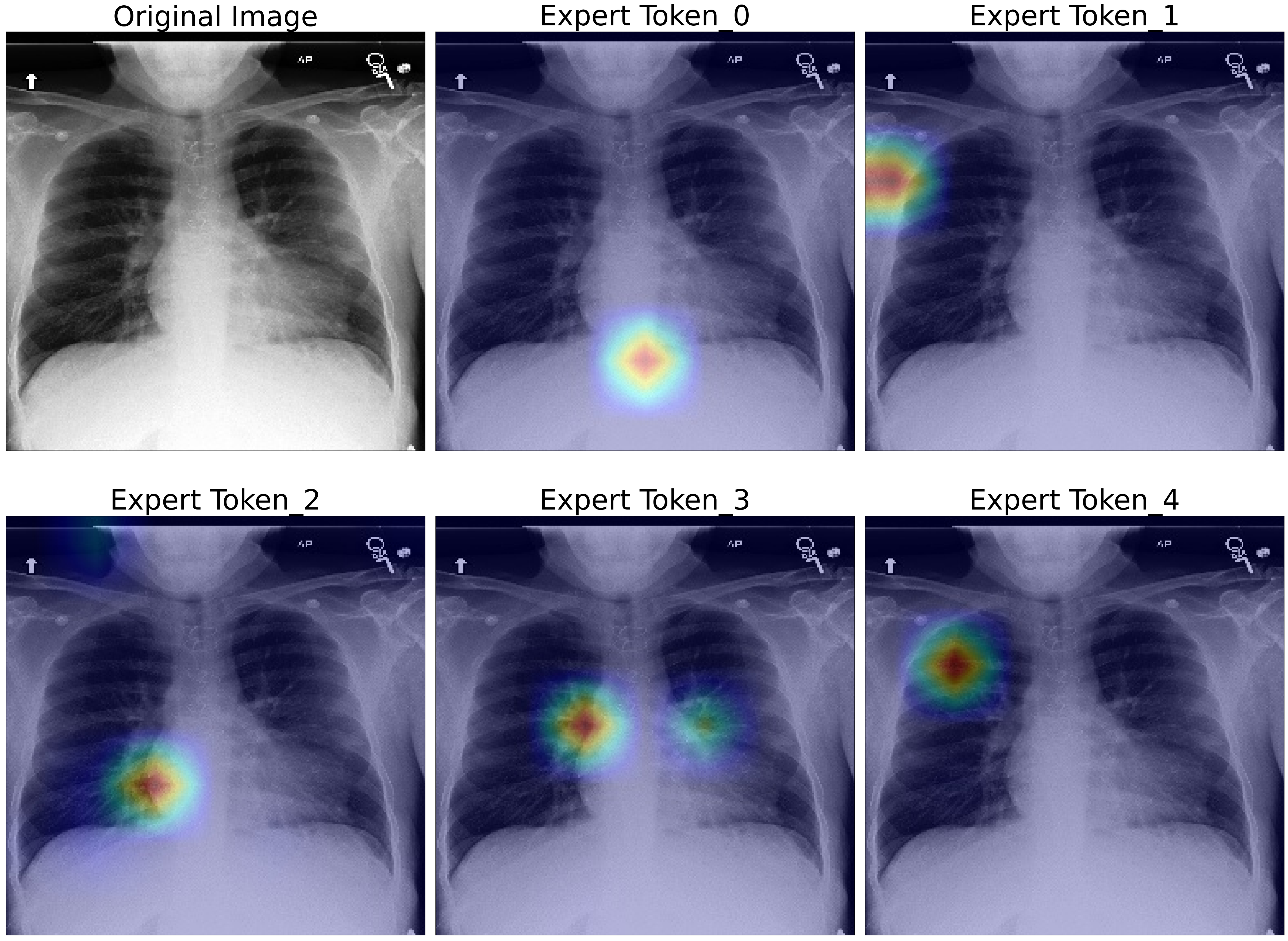}}
      \caption{Attention visualization of expert tokens on the image.}
    \label{fig:visul_att}
    \vspace{-5mm}
\end{figure}

\noindent\textbf{Impact of expert tokens.} To show the impacts of the expert tokens, we train METransformer with different numbers of expert tokens, i.e., $num\_expert\in \left \{ 1,3,5,7,9 \right \}$ and the results on IU-Xray and MIMIC-CXR are shown in Figure.~\ref{fig:num_tokens}. We observe the following. First, increasing the expert tokens can significantly improve the overall performance of the model. This validates the effectiveness of our motivation that by focusing on different image regions with multiple experts, the model can learn more diversified information and thus produce more accurate and diverse diagnostic reports. Second, when the number of expert tokens exceeds a threshold, increasing num\_expert is not able to continue promising a better outcome, for example, when num\_expert is increased from 7 to 9, the CIDEr score on the IU-Xray dataset decreases. A possible explanation is that our model forces each expert token to focus on a different image region (via orthogonal loss), which may cause some experts to attend to irrelevant image regions when there are too many experts, thus may negatively affect the generation process. 

\noindent\textbf{Comparison with ensemble models.} By using multiple experts, our model is conceptually analogous to an ensemble model. We thus compare with three ensemble models in Table.~\ref{tab:ensemble}.  ``Random Initialization" trains randomly initialized METransformer(num\_expert=1) model for 7 times, corresponding to random-seed based ensemble. ``Multiple Decoder" trains the model with the encoder of METransformer(num\_expert=1) and 7 randomly initialized METransformer's decoders, corresponding to late fusion. We also compare with a stochastic model averaging ensemble method. All methods ensemble the results using our proposed Expert Voting strategy. As observed, our METransformer(num\_expert=7) performs significantly better with much fewer trainable parameters. We attribute this to the sophisticated interaction between expert tokens through our compact design. Compared with using a single expert token, our method using 7 expert tokens only incurs 0.007M (0.05\textperthousand) extra parameters, demonstrating its scaling ability. 

\subsection{Qualitative analysis.}
\noindent\textbf{Visualisation of expert tokens.} To gain insight, we visualize the image regions mostly attended by each learned expert token in Figure.~\ref{fig:visul_att} via exploring the attention $\mathbf{\hat{\alpha}}_s$ between the learned expert token embeddings $\hat{\mathbf{z}}_L^e$ and the visual token embeddings $\hat{\mathbf{z}}_L^v$: $\mathbf{\hat{\alpha}}_s = \mathrm{Softmax}(\hat{\mathbf{z}}_L^e (\hat{\mathbf{z}}_L^v)^{T})$
Using $\mathrm{Softmax}(\cdot)$, only the mostly attended image regions are shown. As observed,  each expert token attends to a distinct and critical image region. For example, the image region attended by expert Token\_2 is known as the angle of the rib diaphragm which can provide valuable clinic information.

\noindent\textbf{Qualitative results.} We show the generated results of METransformer compared with the Baseline method (``\#1" in Table.~\ref{tab:ablation_study}) in Figure.~\ref{fig:vis_samples}. On the left, we visualize the image-text attention mapping from the last layer of the expert transformer decoder with three key generated medical terms. As observed, METransformer better aligns the locations with the related text. On the right, we present the corresponding generated reports and the ground truth. For better illustration, we differently color sentences containing different medical information. It is observed that METransformer is able to generate descriptions better aligned with that written by radiologists. For example, METransformer can diagnose anomalies in the heart part, while the Baseline model fails. This is consistent with the fact that our model can better attend to the heart part of the image (see the attented area of ``heart" on the left side of the figure). 


\section{Conclusions}
\label{sec:conclusions}
We present an effective approach for radiology report generation from a new perspective orthogonal to existing research efforts in this field. Our approach follows the concept of multi-specialist consultation to improve the quality of generated reports by introducing multiple learnable expert tokens into a transformer-base framework. Despite its promising performance and properties demonstrated in the experiment, our METransformer is still \textbf{limited} in being a basic framework that could be further enhanced by integrating medical domain knowledge, as seen in ``single-expert" based methods, which will be explored in our future work.  


\newpage
{\small
\bibliographystyle{ieee_fullname}
\bibliography{ref}
}

\end{document}